\documentclass[11pt,a4paper]{article}
\usepackage[hyperref]{conll2018}
\usepackage{times}
\usepackage{latexsym}

\usepackage{url}

\aclfinalcopy % Uncomment this line for the final submission

\usepackage{graphicx}

\usepackage{multirow}
\usepackage{booktabs}

\title{Uncovering divergent linguistic information in word embeddings\\ with lessons for intrinsic and extrinsic evaluation}

\author{Mikel Artetxe, Gorka Labaka, I\~nigo Lopez-Gazpio, Eneko Agirre \\
  IXA NLP Group \\
  University of the Basque Country (UPV/EHU) \\
  {\tt \{mikel.artetxe,gorka.labaka,inigo.lopez,e.agirre\}@ehu.eus}  \\}

\date{}

\begin{document}
\maketitle
\begin{abstract}
Following the recent success of word embeddings, it has been argued that there is no such thing as an ideal representation for words, as different models tend to capture divergent and often mutually incompatible aspects like semantics/syntax and similarity/relatedness. In this paper, we show that each embedding model captures more information than directly apparent. A linear transformation that adjusts the similarity order of the model without any external resource can tailor it to achieve better results in those aspects, providing a new perspective on how embeddings encode divergent linguistic information. In addition, we explore the relation between intrinsic and extrinsic evaluation, as the effect of our transformations in downstream tasks is higher for unsupervised systems than for supervised ones.
\end{abstract}

\section{Introduction} \label{sec:introduction}

Word embeddings have recently become a central topic in natural language processing. Several unsupervised methods have been proposed to efficiently train dense vector representations of words \cite{mikolov2013distributed,pennington2014glove,bojanowski2017enriching} and successfully applied in a variety of tasks like parsing \cite{bansal2014tailoring}, topic modeling \cite{batmanghelich2016nonparametric} and document classification \cite{taddy2015document}.

While there is still an active research line to better understand these models from a theoretical perspective \citep{levy2014neural,arora2016latent,gittens2017skipgram}, the fundamental idea behind all of them is to assign a similar vector representation to similar words. For that purpose, most embedding models build upon co-occurrence statistics from large monolingual corpora, following the distributional hypothesis that similar words tend to occur in similar contexts \citep{harris1954distributional}.

Nevertheless, the above argument does not formalize what ``similar words'' means, and it is not entirely clear what kind of relationships an embedding model should capture in practice. For instance, some authors distinguish between genuine similarity\footnote{Also referred to as \textit{functional similarity} or just \textit{similarity}.} (as in \textit{car - automobile}) and relatedness\footnote{Also referred to as \textit{associative similarity}, \textit{topical similarity} or \textit{domain similarity}.} (as in \textit{car - road}) \citep{budanitsky2006evaluating,hill2015simlex}. From another perspective, word similarity could focus on semantics (as in \textit{sing - chant}) or syntax (as in \textit{sing - singing})  \citep{mikolov2013distributed}. We refer to these two aspects as the two axes of similarity with two ends each: the semantics/syntax axis and the similarity/relatedness axis. 

In this paper, we propose a new method to tailor any given set of embeddings towards a specific end in these axes. Our method is inspired by the work on first order and second order co-occurrences \citep{schutze1998automatic}, generalized as a continuous parameter of a linear transformation applied to the embeddings that we call \emph{similarity order}. While there have been several proposals to learn specialized word embeddings \citep{levy2014dependency,kiela2015specializing,bojanowski2017enriching}, previous work explicitly altered the training objective and often relied on external resources like knowledge bases, whereas the proposed method is applied as a post-processing of any pre-trained embedding model and does not require any additional resource. As such, our work shows that standard embedding models are able to encode divergent linguistic information but have limits on how this information is surfaced, and analyzes the implications that this has in both intrinsic evaluation and downstream tasks. This paper makes the following contributions:
\begin{enumerate}
\item We propose a linear transformation with a free parameter that adjusts the performance of word embeddings in the similarity/relatedness and semantics/syntax axes, as measured in word analogy and similarity datasets.
\item We show that the performance of embeddings as used currently is limited by the impossibility of simultaneously surfacing divergent information (e.g. the aforementioned axes). Our method uncovers the fact that embeddings capture more information than what is immediately obvious.
\item We show that standard intrinsic evaluation offers a static and incomplete picture, and complementing it with the proposed method can offer a better understanding of what information an embedding model truly encodes.
\item We show that the effect of our method also carries out to downstream tasks, but its effect is larger in unsupervised systems directly using embedding similarities than in supervised systems using embeddings as input features, as the latter have enough expressive power to learn the optimal transformation themselves.
\end{enumerate}

All in all, our work sheds light in how word embeddings represent divergent linguistic information, analyzes the role that this plays in intrinsic evaluation and downstream tasks, and opens new opportunities for improvement.

The remaining of this paper is organized as follows. We describe our proposed post-processing in Section \ref{sec:proposed_method}. Section \ref{sec:intrinsic_evaluation} and \ref{sec:extrinsic_evaluation} then present the results in intrinsic and extrinsic evaluation, respectively. Section \ref{sec:discussion} discusses the implications of our work
on embedding evaluation and their integration in downstream tasks. Section \ref{sec:related_work} presents the related work, and Section \ref{sec:conclusions} concludes the paper.

\section{Proposed post-processing} \label{sec:proposed_method}

Let $X$ be the matrix of word embeddings in a given language, so that $X_{i*}$ is the embedding of the $i$th word in the vocabulary. Such embeddings are meant to capture the meaning of their corresponding words in such a way that the dot product $sim(i, j) = X_{i*} \cdot X_{j*}$ gives some measure of the similarity between the $i$th and the $j$th word\footnote{Note that the cosine similarity is the dot product of two length normalized vectors.}. Based on this, we can define the similarity matrix $M(X) = XX^T$ so that $sim(i, j) = M(X)_{ij}$.

Inspired by first order and second order co-occurrences \citep{schutze1998automatic}, one can also define a second order similarity measure on top of this (first order) similarity. In second order similarity, the similarity of two words is not assessed in terms of how similar they directly are, but in terms of how their similarity with third words agrees. For instance, even if $i$ and $j$ are not directly similar, they might both be similar to a third word $k$, which would make them more similar in second order similarity, and one could similarly define third, fourth or $n$th order similarity. The idea that we try to exploit next is that some of these similarity orders can be better at capturing different aspects of language as discussed in Section \ref{sec:introduction}.

More formally, we define the second order similarity matrix $M_2(X) = XX^TXX^T$, so that $sim_2(i, j) = M_2(X)_{ij}$. Note that $M_2(X) = M(M(X))$, so second order similarity can be seen as the similarity of the similarities across all words, which is in line with the intuitive definition given above. More generally, we could define the $n$th order similarity matrix as $M_n(X) = (XX^T)^n$, so that $sim_n(i, j) = M_n(X)_{ij}$. We next show that, instead of changing the similarity measure, one can change the word embeddings themselves through a linear transformation so they directly capture this second or $n$th order similarity.

Let $X^TX = Q \Lambda Q^T$ be the eigendecomposition of $X^TX$, so that $\Lambda$ is a positive diagonal matrix whose entries are the eigenvalues of $X^TX$ and $Q$ is an orthogonal matrix with their respective eigenvectors as columns\footnote{Note that these constraints hold because $X^TX$ is a real symmetric matrix by definition.}. We define the linear transformation matrix $W = Q \sqrt{\Lambda}$ and apply it to the original embeddings $X$, obtaining the transformed embeddings $X' = XW$. As it can be trivially seen, $M(X') = M_2(X)$, that is, such transformed embeddings capture the second order similarity as defined for the original embeddings.

More generally, we can define $W_\alpha = Q \Lambda^\alpha$, where $\alpha$ is a parameter of the transformation that adjusts the desired similarity order. Following the above definitions, such transformation would lead to first order similarity as defined for the original embeddings when $\alpha = 0$, second order similarity when $\alpha = 0.5$ and, in general, $n$th order similarity when $\alpha = (n-1)/2$, that is, $M(XW_0) = M(X)$, $M(XW_{0.5}) = M_2(X)$ and $M(XW_{(n-1)/2}) = M_n(X)$.

Note that the proposed transformation is relative in nature (i.e. it does not make any assumption on the similarity order captured by the embeddings it is applied to) and, as such, negative values of $\alpha$ can also be used to reduce the similarity order. For instance, let $X$ be the second order transformed embeddings of some original embeddings $Z$, so $X = ZW_{0.5}$, where $W_{0.5}$ was computed over $Z$. It can be easily verified that $W_{-0.25}$, as computed over $X$, would recover back the original embeddings, that is, $M(XW_{-0.25}) = M(Z)$. In other words, assuming that the embeddings $X$ capture some second order similarity, it is possible to transform them so that they capture the corresponding first order similarity, and one can easily generalize this to higher order similarities by simply using smaller values of $\alpha$.

All in all, this means that the parameter $\alpha$ can be used to either increase or decrease the similarity order that we want our embeddings to capture. Moreover, even if the similarity order is intuitively defined as a discrete value, the parameter $\alpha$ is continuous, meaning that the transformation can be smoothly adjusted to the desired level.

\begin{table*}[t]
\begin{center}
  \begin{tabular}{clllllll}
    \toprule
    & & \phantom{-} & \multicolumn{2}{c}{Word analogy} & \phantom{-} & \multicolumn{2}{c}{Word similarity} \\
    \cmidrule{4-5} \cmidrule{7-8}
    & & & \multicolumn{1}{c}{Semantic} & \multicolumn{1}{c}{Syntactic} & & \multicolumn{1}{c}{Similarity} & \multicolumn{1}{c}{Relatedness} \\
    & & & & & & \multicolumn{1}{c}{(SimLex-999)} & \multicolumn{1}{c}{(MEN)} \\
    \midrule
    \multirow{2}{*}{word2vec} & Original & & 76.49 & 74.87 & & 44.21 & 76.96 \\
    & Best & & 81.00 {\small $\alpha$ = -0.65} & 74.96 {\small $\alpha$ = 0.10} & & 47.81 {\small $\alpha$ = -0.70} & 78.09 {\small $\alpha$ = -0.30} \\
    \midrule
    \multirow{2}{*}{glove} & Original & & 83.17 & 76.19 & & 40.70 & 80.06 \\
    & Best & & 86.73 {\small $\alpha$ = -0.85} & 76.51 {\small $\alpha$ = -0.10} & & 51.54 {\small $\alpha$ = -0.85} & 84.00 {\small $\alpha$ = -0.45} \\
    \midrule
    \multirow{2}{*}{fasttext} & Original & & 89.76 & 82.44 & & 50.48 & 83.55 \\
    & Best & & 90.85 {\small $\alpha$ = -0.45} & 84.45 {\small $\alpha$ = 0.25} & & 51.55 {\small $\alpha$ = -0.25} & 84.06 {\small $\alpha$ = -0.15} \\
    \bottomrule
  \end{tabular}
\end{center}
\caption{Results in intrinsic evaluation for the original embeddings and the best post-processed model with the corresponding value of $\alpha$. The evaluation measure is accuracy for word analogy and Spearman correlation for word similarity.} \label{tab:results}
\end{table*}

\section{Intrinsic evaluation} \label{sec:intrinsic_evaluation}

In order to better understand the effect of the proposed post-processing in the two similarity axes introduced in Section \ref{sec:introduction}, we adopt the widely used word analogy and word similarity tasks, which offer specific benchmarks for semantics/syntax and similarity/relatedness, respectively.

More concretely, \textbf{word analogy} measures the accuracy in answering questions like ``what is the word that is similar to \textit{France} in the same sense as \textit{Berlin} is similar to \textit{Germany}?'' (semantic analogy) or ``what is the word that is similar to \textit{small} in the same sense as \textit{biggest} is similar to \textit{big}?'' (syntactic analogy) using simple word vector arithmetic \cite{mikolov2013distributed}. The analogy resolution method is commonly formalized in terms of vector additions and subtractions. \citet{levy2014linguistic} showed that this was equivalent to searching for a word that maximizes a linear combination of three pairwise word similarities, so the proposed post-processing has a direct effect on it. For these experiments, we use the dataset published as part of word2vec\footnote{\url{https://github.com/tmikolov/word2vec/blob/master/questions-words.txt}}, which consists of 8,869 semantic and 10,675 syntactic questions of this type \cite{mikolov2013distributed}.

On the other hand, \textbf{word similarity} measures the correlation\footnote{Following common practice, we report Spearman.} between the similarity scores produced by a model and a gold standard created by human annotators for a given set of word pairs.  As discussed before, there is not a single definition of what human similarity scores should capture, which has lead to a distinction between genuine similarity datasets and relatedness datasets. In order to better understand the effect of our post-processing in each case, we conduct our experiments in SimLex-999 \citep{hill2015simlex}, a genuine similarity dataset that consists of 999 word pairs, and MEN \citep{bruni2012distributional}, a relatedness dataset that consists of 3,000 word pairs\footnote{These datasets were selected because the instructions used to elicit human scores are clearly geared towards genuine similarity and relatedness, respectively, and because they have been already used in similar studies \citep{kiela2015specializing}}. 

So as to make our evaluation more robust, we run the above experiments for three popular \textbf{embedding methods}, using large pre-trained models released by their respective authors as follows:

\noindent \textbf{Word2vec} \citep{mikolov2013distributed} is the original implementation of the CBOW and skip-gram architectures that popularized neural word embeddings. We use the pre-trained model published in the project homepage\footnote{\url{https://code.google.com/archive/p/word2vec/}}, which was trained on about 100 billion words of the Google News dataset and consists of 300-dimensional vectors for 3 million words and phrases.

\noindent \textbf{Glove} \citep{pennington2014glove} is a global log-bilinear regression model to train word embeddings designed to explicitly enforce the model properties needed to solve word analogies. We use the largest pre-trained model published by the authors\footnote{\url{http://nlp.stanford.edu/data/glove.840B.300d.zip}}, which was trained on 840 billion words of the Common Crawl corpus and contains 300-dimensional vectors for 2.2 million words.

\noindent \textbf{Fasttext} \citep{bojanowski2017enriching} is an extension of the skip-gram model implemented by word2vec that enriches the embeddings with subword information using bags of character n-grams. We use the largest pre-trained model published in the project website\footnote{\url{https://fasttext.cc/docs/en/english-vectors.html}}, which was trained on 600 billion tokens of the Common Crawl corpus and contains 300-dimensional vectors for 2 million words.

Given that the above models were trained in very large corpora and have an unusually large \textbf{vocabulary}, we decide to restrict its size to the most frequent 200,000 words in each case, leaving the few resulting out-of-vocabularies outside evaluation. In all the cases, we test the proposed post-processing for all the values of the \textbf{parameter $\alpha$} in the $[-1,1]$ range in increments of 0.05. As the goal of this paper is not to set the state-of-the-art but to perform an empirical exploration, we report results across all parameter values on test data.

\subsection{Results on word analogy} \label{subsec:analogy}

\begin{figure*}[t] \centering
\includegraphics[width=\textwidth]{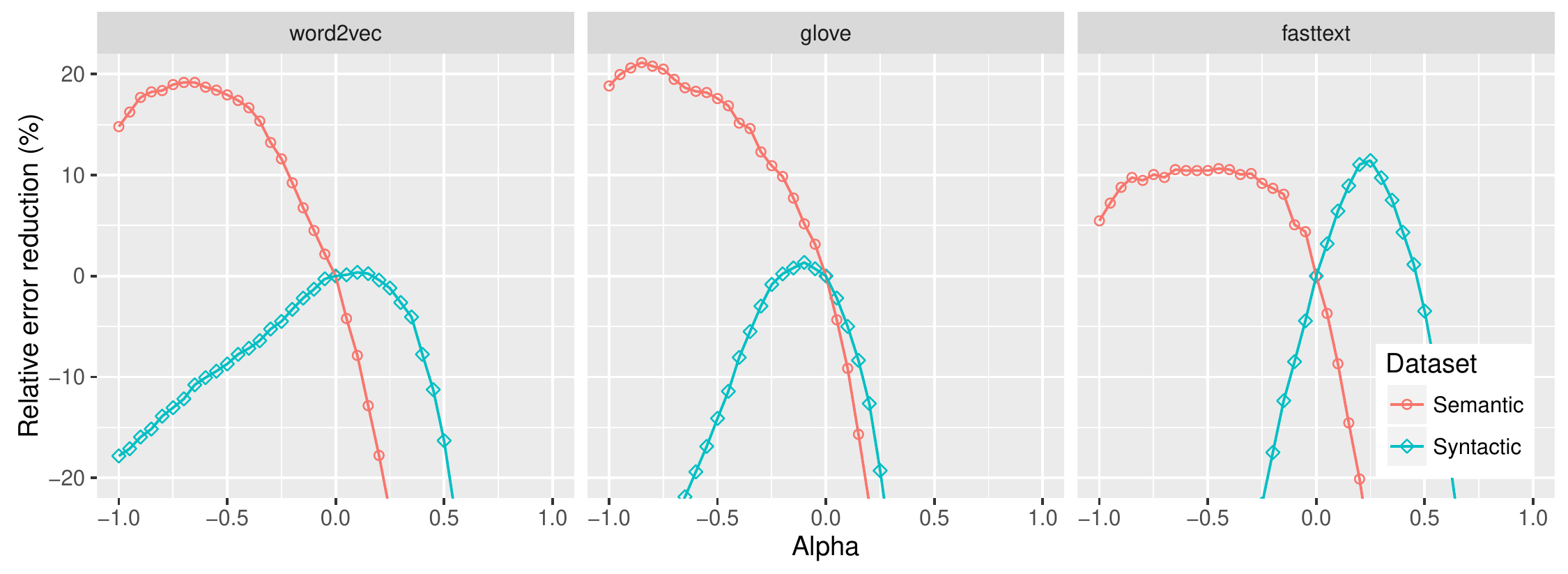}
\caption{Results in word analogy as the relative error reduction with respect to the original embeddings ($\alpha$=0) for different values of $\alpha$.}
\label{fig:analogy}
\end{figure*}

\begin{figure*}[t] \centering
\includegraphics[width=\textwidth]{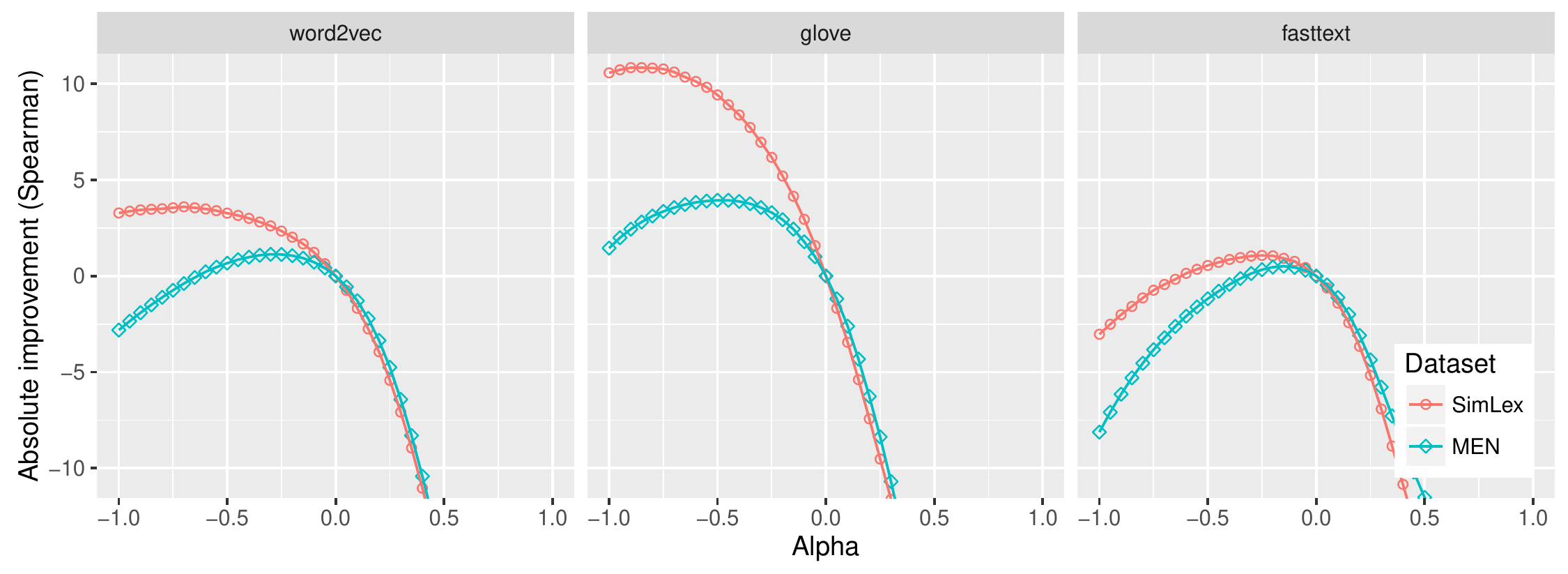}
\caption{Results in word similarity as the absolute improvement in Spearman correlation with respect to the original embeddings ($\alpha$=0) for different  $\alpha$. SimLex for genuine similarity, MEN for relatedness.}
\label{fig:similarity}
\end{figure*}

Table \ref{tab:results} shows the results of the original embeddings ($\alpha=0$) and those of the best $\alpha$, while Figure \ref{fig:analogy} shows the relative error reduction with respect to the original embeddings for all $\alpha$ values\footnote{We choose to show relative error reduction in order to have all curves in the same scale for easier illustration.}. As it can be seen, the proposed post-processing brings big improvements in word analogy, with a relative error reduction of about 20\% in semantic analogies for word2vec and glove and a relative error reduction of about 10\% in both semantic and syntactic analogies for fasttext.

The graphs in Figure \ref{fig:analogy} clearly reflect that, within certain limits, smaller values of $\alpha$ (i.e. lower similarity orders) tend to favor semantic analogies, whereas larger values (i.e. higher similarity orders) tend to favor syntactic analogies. In this regard, both objectives seem mutually incompatible, in that every improvement in one type of analogy comes at a cost of a degradation in the other type. This suggests that standard embedding models already encode enough information to perform better than they do in word analogy resolution, yet this potential performance is limited by the impossibility to optimize for both semantic and syntactic analogies at the same time.

Apart from that, the results also show that, while the general trend is the same for all embedding models, their axes seem to be centered at different points. This is clearly reflected in the optimal values of $\alpha$ for semantic and syntactic analogies (-0.65 and 0.10 for word2vec, -0.85 and -0.10 for glove, and -0.45 and 0.25 for fasttext): the distance between them is very similar in all cases (either 0.70 or 0.75), yet they are centered at different points. This suggests that different embedding models capture a different similarity order and, therefore, obtain a different balance between semantic and syntactic information in the original setting ($\alpha=0$), yet our method is able to adjust it to the desired level in a post-processing step.

\subsection{Results on word similarity} \label{subsec:similarity}

As the results in Table \ref{tab:results} and Figure \ref{fig:similarity} show, the proposed post-processing can bring big improvements in word similarity as well, although there are important differences among the different embedding models tested. This way, we achieve an improvement of about 11 and 4 points for SimLex-999 and MEN in the case of glove, and only 1 and 0.5 points in the case of fasttext, while word2vec is somewhat in between with 3.5 and 1 points.

Following the discussion in Section \ref{subsec:analogy}, this behavior seems clearly connected with the differences in the default similarity order captured by different embedding models. In fact, the optimal values of $\alpha$ reflect the same trend observed for word analogy, with glove having the smallest values with -0.85 and -0.45, followed by word2vec with -0.70 and -0.30, and fasttext with -0.25 and -0.15. Moreover, the effect of this phenomenon is more dramatic in this case: fasttext achieves significantly better results than glove for the original embeddings (a difference of nearly 10 and 3.5 points for SimLex-999 and MEN, respectively), but this proves to be an illusion after adjusting the similarity order with our post-processing, as both models get practically the same results with differences below 0.1 points.

At the same time, although less pronounced than with semantic/syntactic analogies\footnote{Agreeing with the fact that relatedness subsumes similarity \citep{budanitsky2006evaluating}}, the results show clear differences in the optimal configurations for genuine similarity (SimLex-999) and relatedness (MEN), with smaller values of $\alpha$ (i.e. lower similarity levels) favoring the former.

\section{Extrinsic evaluation} \label{sec:extrinsic_evaluation}

\begin{table*}[t]
\begin{center}
  \begin{tabular}{cllll}
    \toprule
    & & & \multicolumn{1}{c}{Centroid} & \multicolumn{1}{c}{DAM} \\
    \midrule
    \multirow{2}{*}{word2vec} & Original & & 65.77 & 72.65 \\
    & Best & & 66.43 {\small $\alpha$ = -0.30} & 73.08 {\small $\alpha$ = 0.10} \\
    \midrule
    \multirow{2}{*}{glove} & Original & & 64.54 & 74.89 \\
    & Best & & 68.96 {\small $\alpha$ = -0.50} & 76.36 {\small $\alpha$ = -0.70} \\
    \midrule
    \multirow{2}{*}{fasttext} & Original & & 69.84 & 77.33 \\
    & Best & & 70.74 {\small $\alpha$ = -0.20} & 77.33 {\small $\alpha$ = 0.00} \\
    \bottomrule
  \end{tabular}
\end{center}
\caption{Results in semantic textual similarity as measured by Pearson correlation for the original embeddings and the best post-processed model with the corresponding value of $\alpha$. The DAM scores are averaged across 10 runs.} \label{tab:sts}
\end{table*}

\begin{figure*}[t] \centering
\includegraphics[width=\textwidth]{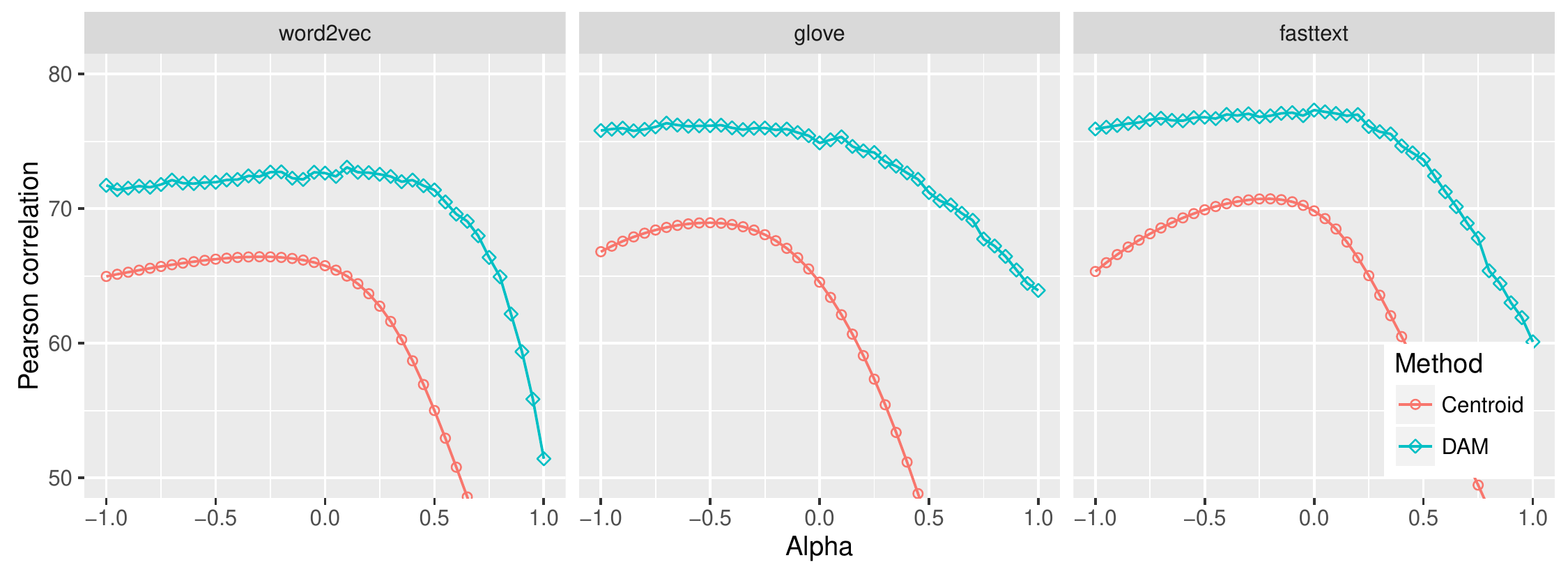}
\caption{Results in semantic textual similarity for different values of $\alpha$. The DAM scores are averaged across 10 runs.}
\label{fig:sts}
\end{figure*}

In order to better understand the effect of the proposed post-processing in downstream systems, we adopt the STS Benchmark dataset on semantic textual similarity \citep{cer2017semeval}\footnote{\url{http://ixa2.si.ehu.es/stswiki/index.php/STSbenchmark}}. This task is akin to word similarity, but instead of assessing the similarity of individual word pairs, it is the similarity of entire sentence pairs as scored by the model that is compared against the gold standard produced by human annotators\footnote{Following common practice, we report Pearson.}. This evaluation is attractive for our purposes because, while the state-of-the-art systems are supervised and based on elaborated deep learning or feature engineering approaches, simpler embedding-based unsupervised models are also highly competitive, making it easier to analyze the effect of the proposed post-processing when integrating the embeddings in a larger model. This way, we test two such systems in our experiments: a simple embedding-based model that computes the cosine similarity between the centroids of each sentence after discarding stopwords, and the Decomposable Attention Model (DAM) proposed by \citet{parikh2016decomposable} and minimally adapted for the task\footnote{\url{https://github.com/lgazpio/DAM_STS}}. The centroid model is thus a simple but very competitive baseline system where the proposed post-processing has a direct effect, whereas DAM is a prototypical deep learning model that uses fixed pre-trained embeddings as input features, producing results that are almost at par with the state-of-the-art in the task.

As the results in Table \ref{tab:sts} and Figure \ref{fig:sts} show, the centroid method is much more sensitive to the proposed post-processing than DAM. More concretely, negative values of $\alpha$ are beneficial for the centroid method up to certain point, bringing an improvement of nearly 4.5 points for glove, and the results clearly start degrading after that ceiling. In contrast, DAM is almost unaffected by negative values of $\alpha$. Positive values do have a clear negative effect in both cases, but the centroid method is much more severely affected than DAM. For instance, for glove, the performance of the centroid method drops 18.19 points when $\alpha=0.50$, in contrast with only 3.69 points for DAM.

This behavior can be theoretically explained by the fact that the proposed post-processing consists in a linear transformation. More concretely, DAM also applies a linear transformation to the input embeddings and, given that the product of two linear transformations is just another linear transformation, its global optimum is unaffected by the linear transformation previously applied by our method. Note, moreover, that the same rationale applies to the majority of machine learning systems that use pre-trained embeddings as input features, including both linear and deep learning models. While there are many practical aspects that can interfere with this theoretical reasoning (e.g. regularization, the optional length normalization of embeddings, the resulting difficulty of the optimization problem...), and explain the variations observed in our experiments, this shows that typical downstream systems are able to adjust the similarity order themselves.

\section{Discussion} \label{sec:discussion}

Our experiments reveal that standard word embeddings encode more information than what is immediately obvious, yet their potential performance is limited by the impossibility of optimally surfacing divergent linguistic information at the same time. This can be clearly seen in the word analogy experiments in Section \ref{subsec:analogy}, where we are able to achieve significant improvements over the original embeddings, yet every improvement in semantic analogies comes at the cost of a degradation in syntactic analogies and vice versa. At the same time, our work shows that the effect of this phenomenon is different for unsupervised systems that directly use embedding similarities and supervised systems that use pre-trained embeddings as features, as the latter have enough expressive power to learn the optimal balance themselves.

We argue that our work thus offers a new perspective on how embeddings encode divergent linguistic information and its relation with intrinsic and extrinsic evaluation as follows:

\begin{itemize}
\item Standard intrinsic evaluation offers a static and incomplete picture of the information encoded by different embedding models. This can be clearly seen in the word similarity experiments in Section \ref{subsec:similarity}, where fasttext achieves significantly better results than glove for the original embeddings, yet the results for their best post-processed embeddings are at par. As a consequence, if one simply looks at the results of the original embeddings, they might wrongly conclude that fasttext is vastly superior to glove at encoding semantic similarity information, but this proves to be a mere illusion after applying our post-processing. As such, intrinsic evaluation combined with our post-processing provides a more complete and dynamic picture of the information that is truly encoded by different embedding models.
\item Supervised systems that use pre-trained embeddings as features have enough expressive power to learn the optimal similarity order for the task in question. While there are practical aspects that interfere with this theoretical consideration, our experiments confirm that the proposed post-processing has a considerably smaller effect in a prototypical deep learning system. This reinforces the previous point that standard intrinsic evaluation offers an incomplete picture, as it is severely influenced by an aspect that has a much smaller effect in typical downstream systems. For that reason, using our proposed post-processing to complement intrinsic evaluation offers a better assessment of how each embedding model might perform in a downstream task.
\item Related to the previous point, while our work shows that the default similarity order captured by embeddings has a relatively small effect in larger learning systems as they are typically used, this is not necessarily the best possible integration strategy. If one believes that a certain similarity order is likely to better suit a particular downstream task, it would be possible to design integration strategies that encourage it to be so during training, and we believe that this is a very interesting research direction to explore in the future. For instance, one could design regularization methods that penalize large deviations from this predefined similarity order.
\end{itemize}

\section{Related work} \label{sec:related_work}

There have been several proposals to learn word embeddings that are specialized in certain linguistic aspects. For instance, \citet{kiela2015specializing} use a joint-learning approach and two variants of retrofitting \citep{faruqui2015retrofitting} to specialize word embeddings for either similarity or relatedness. At the same time, \citet{levy2014dependency} propose a modification of skip-gram that uses a dependency-based context instead of a sliding windows, which produces embeddings that are more tailored towards genuine similarity than relatedness. \citet{bansal2014tailoring} follow a similar approach to train specialized embeddings that are used as features for dependency parsing. Finally, \citet{mitchell2015orthogonality} exploit morphology and word order information to learn embeddings that decompose into orthogonal semantic and syntactic subspaces. Note, however, that all these previous methods alter the training objective of specific embedding models and often require additional resources like knowledge bases and syntactic annotations, while the proposed method is a simple post-processing that can be applied to any embedding model and does not require any additional resource.

Other authors have also proposed post-processing methods for word embeddings with different motivations. For instance, \citet{faruqui2015sparse} transform word embeddings into more interpretable sparse representations, obtaining improvements in several benchmark tasks. \citet{rothe2016ultradense} propose an orthogonal transformation to concentrate the information relevant for a task in a lower dimensional subspace, and \citet{rothe2016word} extend this work to decompose embeddings into four subspaces specifically capturing polarity, concreteness, frequency and part-of-speech information. Finally, \citet{labutov2013reembedding} perform unconstrained optimization with proper regularization to specialize embeddings in a supervised task.

The proposed method is also connected to a similar parameter found in traditional count-based distributional models as introduced by \citet{caron2001experiments} and further analyzed by \citet{bullinaria2012extracting} and \citet{turney2012domain}. More concretely, these models work by factorizing some co-occurrence matrix using singular value decomposition, so given the co-occurrence matrix $M=USV^T$, the word vectors will correspond to the first $n$ dimensions of $W=US^\alpha$, where the parameter $\alpha$ plays a similar role as in our method. Note, however, that our proposal is more general and can be applied to any set of word vectors in a post-processing step, including neural embedding models that have superseded these traditional count-based models as we in fact do in this paper.

Finally, there are others authors that have also pointed limitations in the intrinsic evaluation of word embeddings. For instance, \citet{faruqui2016problems} and \citet{batchkarov2016critique} argue that word similarity has many problems like the subjectivity and difficulty of the task, the lack of statistical significance and the inability to account for polysemy, warning that results should be interpreted with care. \citet{chiu2016intrinsic} analyze the correlation between results on word similarity benchmarks and sequence labeling tasks, and conclude that most intrinsic evaluations are poor predictors of downstream performance. In relation to that, our work explains how embeddings encode divergent linguistic information and the different effect this has in intrinsic evaluation and downstream tasks, showing that the proposed post-processing can be easily used together with any intrinsic evaluation benchmark to get a more complete picture of the representations learned.

\section{Conclusions and future work} \label{sec:conclusions}

In this paper, we propose a simple post-processing to tailor word embeddings in the semantics/syntax and similarity/relatedness axes without the need of additional resources. By measuring the effect of our post-processing in word analogy and word similarity, we show that standard embedding models are able to encode more information than what is immediately obvious, yet their potential performance is limited by the impossibility of optimally surfacing divergent linguistic information. We analyze the different role that this phenomenon plays in intrinsic and extrinsic evaluation, concluding that intrinsic evaluation offers a static picture that can be complemented with the proposed post-processing, and prompting for better integration strategies for downstream tasks. We release our implementation at \url{https://github.com/artetxem/uncovec}, which allows to easily reproduce our experiments for any given set of embeddings.

In the future, we would like to explore better integration strategies for machine learning systems that use pre-trained embeddings as features, so that downstream systems can better benefit from previously adjusting the embeddings in the semantics/syntax and similarity/relatedness axes. At the same time, we would like to extend our analysis to more specialized embedding models \citep{kiela2015specializing,levy2014dependency} to get a more complete picture of what information they capture.

\section*{Acknowledgments}

This research was partially supported by the Spanish MINECO (TUNER TIN2015-65308-C5-1-R, MUSTER PCIN-2015-226 and TADEEP TIN2015-70214-P, cofunded by EU FEDER), the UPV/EHU  (excellence research group), and the NVIDIA GPU grant program. Mikel Artetxe and I\~nigo Lopez-Gazpio enjoy a doctoral grant from the Spanish MECD.

\bibliography{conll2018}
\bibliographystyle{acl_natbib_nourl}

\end{document}